\pdfoutput=1

\documentclass[11pt]{article}

\usepackage[final]{acl}

\usepackage{times}
\usepackage{latexsym}

\usepackage[T1]{fontenc}

\usepackage[utf8]{inputenc}

\usepackage{microtype}

\usepackage{inconsolata}

\usepackage{graphicx}
\usepackage{float}  
\usepackage{algorithm}
\usepackage{amsmath}
\usepackage{placeins}
\usepackage{subcaption}
\usepackage{multirow}
\usepackage{hyperref}

\title{Data-driven Coreference-based Ontology Building}

\author{Shir Ashury-Tahan$^{1}$$^{2}$, Amir David Nissan Cohen$^{1}$, Nadav Cohen$^{1}$, \\
\textbf{Yoram Louzoun$^{1}$ \and Yoav Goldberg$^{1}$$^{3}$}\\
$^{1}$Bar-Ilan University, $^{2}$IBM Research, $^{3}$Allen Institute for AI \\
\texttt{\href{mailto:shirashury@gmail.com}{shirashury@gmail.com}}
}

\begin{document}
\maketitle
\begin{abstract}
While coreference resolution is traditionally used as a component in individual document understanding, in this work we take a more global view and explore what can we learn about a domain from the set of all document-level coreference relations that are present in a large corpus. 
We derive coreference chains from a corpus of 30 million biomedical abstracts and construct a graph based on the string phrases within these chains, establishing connections between phrases if they co-occur within the same coreference chain. We then use the graph structure and the betweeness centrality measure to distinguish between edges denoting hierarchy, identity and noise, assign directionality to edges denoting hierarchy, and split nodes (strings) that correspond to multiple distinct concepts. The result is a rich, data-driven ontology over concepts in the biomedical domain, parts of which overlaps significantly with human-authored ontologies. We release the coreference chains and resulting ontology \footnote{\url{https://huggingface.co/spaces/biu-nlp/Data-driven_Coreference-based_Ontology}} under a creative-commons license, along with the code \footnote{\url{https://github.com/ShirApp/Coreference-based-Ontology-Building}}.
\end{abstract}

\begin{figure}

    \centering
    \includegraphics[width=\columnwidth]{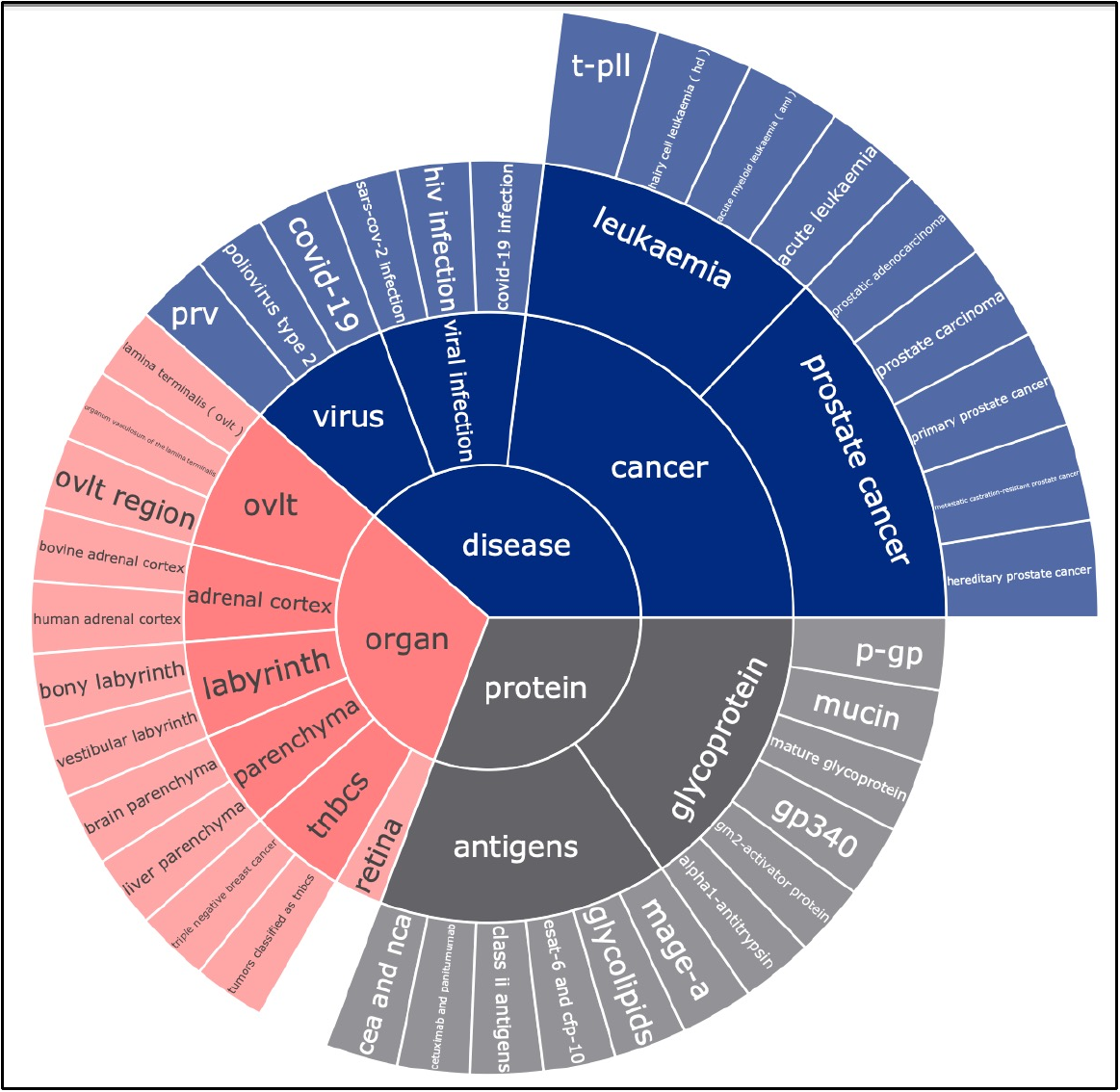}
    \caption{\textbf{Resulting Ontology Example} that may reflect the type of structure achievable using our method. } 
    \label{example_ontology_fig}
\end{figure}

\section{Introduction}

Ontologies categorize concepts into groups and arranges them in a hierarchy and are essential for researchers in the biomedical domains \cite{ontologies1,ontologies2,ontologies3}, as evidenced by the vast number of ontologies available in repositories such as BioPortal\footnote{\url{https://bioportal.bioontology.org/}}. These ontologies are predominately human curated, they each contains a collection of concepts arranged in a hierarchy, and for each concept a list of aliases, which are different equivalent names for this concept. 
While useful, such ontologies have deficiencies: being manually curated they are both expensive to create and maintain and also non-comprehensive; they do not cover all areas of interest a researcher may be interested in, especially for long-tail interests (for example, BioPortal does not contain an ontology containing a comprehensive list of peptides); and the concept names and their aliases may not be aligned with how the concepts appear in text, reducing their utility for text mining applications \cite{Blair2014QuantifyingTI} (for example, the UMLS ontology entry for ``Alzheimer's disease'' does not contain the string ``alzheimer'', although it is a very common way to refer to this condition in text). 
Thus, a data-driven, text-based ontology derived directly from the scientific literature can be of immense value: (a) it will provide coverage of all (or most) the concepts that appear in the text, including long-tail ones, arranged in hierarchies based on their actual use in scientific texts; (b) concepts names and aliases will be naturally aligned with their text appearances; and (c) they can aid manual creation, extension and maintenance of existing ontologies by surfacing areas of deficiencies in coverage, and suggesting alternative hierarchies and potentially missing entries.

In this work we propose to create such a data-driven ontology from text, using a novel signal: the topology of a graph created by running in-document coreference resolution over scientific documents, and creating a graph where nodes are textual strings and edges represent that the two strings participated in a coreference chain. We show that his graph's topology contains rich information which allows to identify concepts, aliases and hierarchies like the ones in the Figure \ref{example_ontology_fig}. 

We exploit the dynamics of phrase co-occurrence within the graph, observing a correlation between a phrase's contribution to information flow and its level of generality. Therefore, our approach centers on a single centrality measure, specifically betweenness centrality \cite{freeman1977set}, aimed at understanding information flow. This measure guides the transformation of the graph into a directed structure, establishing the framework for ontology construction.

\section{Coreference-based Ontology Construction}

\subsection{Coreference Graph Construction}

We run a coreference resolution algorithm \cite{otmazgin-etal-2023-lingmess}\footnote{We selected this algorithm for its highly efficient runtime, with minimal impact on performance, which is crucial for handling large corpora like those used in our experiments.} on each of 30M PubMed abstracts to extract coreference chains from each abstract. Each coreference chain is a list of phrases that occur in the same document, and were determined by the coreference algorithm to co-refer to the same concept. 

We filter phrases that correspond to pronouns and stop words (as determined by SciSpacy \cite{neumann-etal-2019-scispacy} and NLTK \cite{bird2009natural}), remove stop-words, pronouns, determiners and quantifiers from the beginning of phrases, and unified singular and plural versions of phrases. We further remove phrases which we determine to contain only verbs, as these stem mostly from coreference mistakes, and do not correspond to entities. We then designate each of the unique remaining phrases as nodes, and connect two nodes if their phrases co-occur in the same coreference chain, weighing the edge by the number of chains in which this pair co-occurs. (The same phrase may appear in different documents, hence participating in multiple chains)

The resulting graph $G$ has over $3$ million nodes and approximately $7$ million weighted edges.

\subsection{Ontology Extraction}
Our aim is to take the corefence graph $G$ and extract an ontology: a directed acyclic graph where each node corresponds to a concept, and an edge from node B to node A indicates that A is more specific than B (``A is a B''). Each node is associated with one or more strings which are aliases for this concept. To extract an ontology from $G$, we:

\begin{enumerate}
    \item Identify equivalence relations between nodes, which will form the aliases. We do this by marking some edges in $G$ as indicating \emph{identity}.
    \item Mark the remaining edges in $G$ as indicating a hierarchy, and assign them a direction.
    \item Split some nodes where the same string corresponds to multiple distinct concepts.
    \item Tag some edges in $G$ as noisy or irrelevant.
\end{enumerate}
  
At a high level, we utilize estimated betweenness centrality values of the nodes to determine the kind and direction of each edge in the graph, thereby transforming the graph into a Directed Acyclic Graph (DAG), from which we will derive the ontology. 

\paragraph{Betweenness Centrality as a Main Measure}
The coreference graph is undirected, and we wish to assign edges with direction that indicate IS-A relations. 
A major observation is that phrases that denote concepts that are higher-up in the IS-A hierarchy (are more general) co-occur in many different coreference clusters, and with many different phrases, while phrases that are more specific belong in only few clusters, with a more restricted set of phrases (e.g., concepts like "disease" will appear in many clusters denoting specific diseases, "lung diseases" will appear with "disease" as well as with many specific lung diseases, while "asthma" may share a cluster with "diseases" and with "lung disease", but likely not with other lung diseases). Consequently, if we choose two random nodes in the graph, if there is a path between them it either goes directly from the less specific to the more specific one, or it goes to a common shared parent (which is more general than both) and then back. Thus, we expect the more general concepts to be on more paths connecting pairs of nodes in the graph. This is precisely the notion that is captured by the \emph{betweeness centrality} measure. 

Therefore, the first step in establishing the edge direction is to compute the betweeness centrality score of each node, and assign the direction of an edge to be from the the node with higher centrality (more general nodes) to one with lower centrality. When both nodes havea centrality score of zero (these nodes don't connect different concepts) we denote them as identity edges.

As exact centrality computation is expensive---$O(V^2 + V \times E)$ when
using the fast algorithm of \citet{brandes2001faster}---and our graph is large, we opted for an approximate solution that relies on performing a restricted number of shortest-path computations above a small set of randomly chosen pivots \cite{brandes2007centrality}. Using this approximation with $500$ pivots works well for our purpose without reducing accuracy, as we use centrality scores only for computing relative ordering, rather than needing the exact values (as also detailed in Appendix \ref{appendix:betweeness}).

\paragraph{From Graph to DAG} \label{subsection_graph_to_dag}

Roughly $70\%$ of the nodes have a betweenness value of $0$, suggesting they function as leaves in the DAG hierarchy. This is while $240{,}000$ of the edges in the graph connect such nodes. Consulting a random sample of edges reveal that they indeed connect aliases of the same concept. We subsequently mark these edges as (unordered) identity edges, indicating alises. The rest of the edges ($6.7$ million edges) are marked temporarily as indicating a hierarchy, and we assign direction from the node with higher betweeness score to the one with the lower score. (In the next steps, we will reveal that some of these "hierarchy edges" are actually identity edges or noise, and we relabel them as such.)

\paragraph{Betweenness Centrality Challenges}

Using betweenness centrality for determining edge direction is sensitive to common nodes that represent specific entities. These nodes often exhibit higher betweenness than their neighbors, leading to misleading hierarchical relations. We identified three instances of this:

\begin{enumerate}
    \item A node that co-occurs with other phrases in coreference chains more frequently than its more general neighbor (e.g., "COVID-19" vs. "epidemic"), as illustrated in Figure \ref{fig_case1}.
    \item A specific entity name appears more frequently than its aliases (e.g., "gys2" vs. "glycogen synthase 2"), as depicted in Figure \ref{fig_case2}.
    \item A node shares a common name with multiple entities (e.g., "IL" for "Illinois," "IL-6," "IL-8"), as shown in Figure \ref{fig_case3}.
\end{enumerate}

The first case leads to incorrect hierarchy direction, while the second leads to incorrect tags. 

For the first case, we establish that names are more specific than general nouns, so we change the direction of edges from nouns to names, correcting around 200,000 edges\footnote{Names are identified based on capitalization; consistently capitalized terms are names, while others are nouns.}.


The graph structure alone often fails to distinguish between concepts, as seen in cases 2 and 3. In case 2, all child nodes represent the same entity as the parent but rarely co-occur, making their identity relationship hard to detect. In case 3, an ambiguous string refers to multiple entities, leading child nodes to represent distinct entities. The lack of connections may result from non-co-occurrence or indicate separate entities. To resolve this, we identify child nodes as case 2 or 3, analyze their semantic context in a separate subgraph, connect nodes to their k-nearest neighbors, and cluster them into distinct concepts or confirm they represent a single entity. We expect the semantic representations to help establish strong connections within similar nodes, while maintaining sparse connectivity between distant nodes, with clustering used to extract the strongly connected groups.


We designed Algorithm \ref{alg:names_hierarchical} to address these cases by grouping strings into distinct entities. We embed the node $n$ and its children $C = {c_1, c_2, \ldots, c_l}$ using a language model\footnote{We used the S-BERT model \cite{Reimers2019SentenceBERTSE} from \cite{deka2022improved}, trained on health data. The only input to the model was the phrase itself.}, capturing their semantic similarities. Next, we create a KNN subgraph, where $n$ and $c_i \in C$ are the nodes, and each is connected by an edge to its $k$ closest neighbors. 
The Louvain algorithm \cite{Blondel_2008} is then applied to the subgraph to detect communities (dense areas) representing distinct senses, allowing us to split nodes accordingly, if necessary. Finally, each community is treated as a concept. If one community is detected (case 2), we merge $n$ and $c_i \in C$ into a single concept. Otherwise (case 3), each detected community is treated as a separate concept, and $n$ is split into different nodes (with the same string) for each concept. During this process we tagged about $230{,}000$ edges from hierarchical to identity.

\begin{algorithm*}[t]
\caption{Split and Classify Names with Hierarchical Behavior to Senses Algorithm}\label{alg:names_hierarchical}
\begin{minipage}{\textwidth}
\begin{enumerate}
    \item Identify all nodes that are names (starting with a capital letter) and are hierarchical in our DAG (have children).
    \item For each node $n$, and its children $C=\{c_1, c_2, \ldots , c_l\}$:
    \begin{enumerate}
        \item Embed each string in $V=\{n\} \cup C$ using an LLM.
        \item Create a nearest neighbour graph over $V$, where each node has an edge to its $k$ closest nodes.
        \item Use the Louvain algorithm to find the communities, representing the senses.
        \item Split $n$ into $|communities|$ nodes and unify each split with a community. The split node would inherit only the common parents with the concept it is connected to.
    \end{enumerate}
\end{enumerate}
\end{minipage}
\end{algorithm*}

\paragraph{Cleaning noisy edges}

We observe unwanted noisy edges in the graph that connect very common phrases (e.g. "group" and "variant") that are not supposed to be connected. 
These edges arise from pairs of phrases that are mistakenly assigned to the same coreference cluster.

In these cases, the erroneous relations are due to mistakes made by the coreference annotator, and their edges have much lower weight compared to other relations in which their respective nodes participate. We therefore used the PMI measure \cite{fano1961transmission} to calculate the association between each pair of connected phrases. Edges whose association was less than expected by chance, i.e., those with negative PMI values, were filtered out.

Let the probability \( P(\text{phrase}_i) \) be defined as the ratio of the count of co-occurrences of phrase \( \text{phrase}_i \) with other phrases to the total number of co-occurrences of all phrases in the corpus: $P(\text{phrase}_i) = \frac{\text{count}(\text{phrase}_i)}{\sum_{k=1}^{N} \text{count}(\text{phrase}_k)}$
where \( N \) is the total number of distinct phrases in the corpus. Let us also denote the joint probability \( P(\text{phrase}_i, \text{phrase}_j) \), which is defined as the number of co-occurrences of phrases \( \text{phrase}_i \) and \( \text{phrase}_j \) divided by the total number of co-occurrences in the corpus: 
$P(\text{phrase}_i, \text{phrase}_j) = \frac{\text{count}(\text{phrase}_i, \text{phrase}_j)}{\sum_{k=1}^{N} \text{count}(\text{phrase}_k)}$
We calculated the PMI for each pair of phrases connected by an edge in the graph as follows: $PMI(\text{phrase}_i, \text{phrase}_j) = \log \left( \frac{P(\text{phrase}_i, \text{phrase}_j)}{P(\text{phrase}_i) \cdot P(\text{phrase}_j)} \right)$. We identified approximately $350{,}000$ such edges and labeled them as noise.

\paragraph{The Final Graph} 
Overall, we were able to label over $6$ million graph edges. We marked most of them as indicating an identity or hierarchy relation, and the rest as noise. We found the hierarchical relation to be much more common in our graph. We marked approximately $5.3$ million edges as directed edges indicating a hierarchy, and about $440{,}000$ as identity edges. The rest of $350{,}000$ edges are tagged as noise.

\section{Evaluation and Results}
Evaluating the quality of the resulting graph is challenging, as there is no ground-truth to compare to \cite{McCrae2009AutomaticEO}. Still, we compare our results to existing human curated ontologies in the biomedical domain (UMLS \cite{bodenreider2004umls} and SnomedCT \cite{loinc-585}), and assess how well we manage to capture concepts from them. UMLS provides aliases for identity nodes, while SnomedCT provides hierarchical relations and directions between concepts. If these resources were perfect, we wouldn't need to create the data-driven one to begin with. We thus combine automatic metrics with human evaluations.

\paragraph{Evaluating Hierarchy Assignments}
We compare ourselves to SnomedCT, an ontology with 1.4M medical phrases and 1.7M corresponding "is a" relation tuples. We consider only edges between the strings that are available in both SnomedCT and our data resulting in $226{,}278$ edges for evaluation. Let \emph{correct} denote the number of predicted hierarchy edges that participate in the same hierarchy in SnomedCT (there is a directed path between them in SnomedCT). We compute \emph{precision} as \emph{correct / all predicted hierarchy edges}, and \emph{recall} as \emph{correct / all edges that are marked as hierarchy in SnomedCT}.
We achieve a high \textbf{recall of 84.3\%}, with a \textbf{lower precision, at only 40.1\%}.
However, examining the precision error reveals that many cases stem from valid disagreements between the resources. For example, our graph places "\emph{defibrillation}" under "\emph{procedure}", which is not reflected in SnomedCT.  We thus sample 100 random hierarchy edges and annotate them manually (not compared to SnomedCT), revealing a \textbf{substantially higher precision of 75\%}.

\paragraph{Hierarchy Edge Direction Evaluation} For hierarchy edges whose end-points are reachable also in SnomedCT, we find the edge direction is \textbf{consistent with SnommedCT in 92.1\% of the cases}. 

\paragraph{Identity-edge Evaluation}
Finally, we evaluate the accuracy of the identity edges, which represent aliases. Here, we value precision over recall: it is better to miss an alias than to introduce an incorrect one: mistaking an alias relation for a hierarchical one is not as bad as erroneously equating two concepts. Here, we compare to UMLS aliases, focusing on the $29{,}798$ strings that are shared between our ontology and UMLS. We treat identity edges as inducing clusters, evaluate the clustering using two metrics: entropy, to measure the homogeneity of the predicted clusters compared to a gold standard (lower means more homogeneous) and Adjusted Rand Index (ARI) to measures similarity between our clustering and UMLS's. We obtain an \textbf{entropy of 0.406} for the predicted clusters, suggesting the clusters are reasonably pure (do not contain many erroneous entries). Moreover, the moderate \textbf{ARI score of 0.387} indicates that our clusters are also split well.

The evaluation suggests our approach aligns well with established ontologies. Human assessments of hierarchical edges and entropy metrics indicate near-precision, while recall and ARI measures suggest the extracted ontology is close to complete.

\begin{table}[H]   
\centering
\resizebox{0.95\columnwidth}{!}{ 

\begin{tabular}{|l|l|l|l|}
\hline
\textbf{Ontology} & \textbf{Task}              & \textbf{Measure} & \textbf{Score} \\ \hline
\multirow{3}{*}{SnomedCT} & Hierarchy edges      & Precision        & 0.401          \\ \cline{3-4} 
                          &                     & Recall           & 0.842          \\ \cline{3-4} 
                          &                     & F1               & 0.541          \\ \cline{2-4} 
                          & Hierarchy Directions & Precision        & 0.921          \\ \hline
\multirow{2}{*}{UMLS}     & Identity edges       & Entropy          & 0.406          \\ \cline{3-4} 
                          &                     & ARI              & 0.387          \\ \hline
\end{tabular}
}
\caption{\textbf{Our results compared to other ontologies} show that we successfully identified the majority of hierarchy edges, with a small number of errors in concepts.}
\end{table}

\section{Related Work}

Ontology learning, essential for many applications, has traditionally relied on linguistic approaches, using pattern extraction and syntactic analysis combined with statistical methods. While syntactic patterns can be effective, they often require human intervention. A well-known example is \cite{6287662}, which designed a process to extract patterns for specific relations from phrase pairs.
Some works apply syntactic analysis by parsing dependency trees to extract relations and build ontologies automatically, as seen in \cite{inproceedings,Gamallo2002MappingSD}. 
Statistical methods like \cite{Drymonas2010UnsupervisedOA} use hierarchical clustering for taxonomy building and conditional probabilities for non-taxonomic relations, while \cite{Faure1998ACC} applies conceptual clustering to automatically acquire and organize subcategorization frames.

These techniques differ from ours by focusing on local relations, while we make use of global signals---cross-document-level coreference relations---to identify relationships more globally within the corpus.

Recent studies have explored using large language models (LLMs) for ontology building due to their ability to capture contextual information \cite{Brown2020LanguageMA, Devlin2019BERTPO, Achiam2023GPT4TR}. For example, \cite{BabaeiGiglou2023LLMs4OLLL} applied a zero-shot prompting approach to term typing, taxonomy discovery, and relation extraction, while \cite{Funk2023TowardsOC} used ordered prompts to construct concept hierarchies.
Both studies show that while LLMs can aid ontology learning, they cannot yet construct ontologies independently.

\section{Conclusions}
We demonstrated that a text-based, data-driven biomedical ontology\footnote{\url{https://huggingface.co/spaces/biu-nlp/Data-driven_Coreference-based_Ontology}} can be created by considering the topology of a coreference graph obtained from a large corpus.
Furthermore, we achieved this primarily through the use of a single centrality measure. A significant contribution of this approach is its generality, allowing for easy adaptation to other fields. Additionally, our method is scalable and can be implemented for networks of varying sizes.
Compared to existing ontologies, we obtain very accurate directionality and high recall of hierarchical structure. We also find accurate hierarchical relations that are not reflected in the human-curated ontologies.
In future work, our automatically constructed ontology could be applied to downstream tasks.

\section{Limitations}

\paragraph{Evaluation difficulties.} Assessing our unsupervised approach poses challenges in achieving a comprehensive and scalable evaluation. Direct comparisons to established ontologies are complicated, as these may not fully capture diverse language usages present in extensive corpora. Manual evaluations, limited by scalability, may not be wholly representative of our graph's overall quality.

\paragraph{Error Propagation.} The propagation of errors or inconsistencies from the corpus into the ontology might compromise its quality and accuracy.

\section{Ethics Statement}
We do not identify ethical concerns with this work. The resulting ontology is useful but not perfectly accurate, and must be used with care and using human oversight.

\bibliography{custom}

\clearpage
\appendix


\section{Centrality Approximation for Betweenness in Large Networks} \label{appendix:betweeness}

\subsection{Overview of Betweenness Centrality}

Betweenness centrality \cite{freeman1977set} is a centrality measure that aids, among other things, in the analysis and characterization of graphs. It is particularly valuable in networks where intermediaries play a crucial role in facilitating information flow or identifying connectivity. At its core, betweenness centrality quantifies the significance of a node within a network by assessing how frequently it lies on the shortest paths between other nodes. Mathematically, for a given node \( v \), its betweenness centrality \( BC(v) \) is defined as:

\[
BC(v) = \sum_{s \neq v \neq t} \frac{\sigma_{st}(v)}{\sigma_{st}}
\]

where \( \sigma_{st} \) denotes the total number of shortest paths from node \( s \) to node \( t \), and \( \sigma_{st}(v) \) represents the number of those paths that pass through node \( v \).

However, computing the betweenness centrality for each node in the graph requires calculating all shortest paths between pairs of nodes. The fastest known algorithm for this task \cite{brandes2001faster} has a time complexity of $O(V^2 + V \times E)$.

\subsection{Approximation Algorithm}

As networks scale in size, such as in our case, the number of shortest paths that must be computed is huge, making exact calculations infeasible. A practical solution is to employ an approximate betweenness centrality calculation. This approach is particularly valid when only relative importance is required, as is the case in our work where we aim to establish an order among the nodes.

To address the computational cost, we sought an appropriate approximation that could dramatically reduce the number of computations while still allowing us to create the desired order. The only approximation that is adapted to large network as ours is \cite{brandes2007centrality}, which proposes a solution that relies on performing a limited number of shortest-path computations over a small set of randomly chosen pivots. 
By focusing on this smaller subset, the number of path computations is significantly reduced, yielding a complexity of \( O(k(V+E)) \), where \( k \) represents the number of sampled nodes. Increasing \( k \) enhances accuracy but may also extend computation time. The main constraints in their framework indicate that the number of pivots should be greater than \( \log(V) \) and that the graph's diameter should be constant, conditions that our case satisfies.

\subsection{Experimental Results}

\paragraph{Accuracy of Results.} Since the algorithm assists in creating a DAG, we evaluated its performance based on two criteria. The first criterion is its accuracy in determining the direction of edges in the graph that represent hierarchical relations within the ontologies we compared. The second criterion assesses its accuracy in identifying leaves that are connected to the same concept. We found both metrics to have high scores for \(k=500\): $91.3$ for the direction of edges and $86.7$ for the accuracy of the connected leaves.

\paragraph{Consistency of Results.} To measure approximation consistency and variability, we executed the algorithm \(5\) times on our graph, using different \(k\) values: \(100, 500, 1{,}000, 2{,}000, 2{,}500\). We calculated the order for each edge in each run and discovered that only \(6\%\) of all edges were conflicted (at least one run disagreed with the others regarding the edge direction). Furthermore, only \(0.01\%\) of the cases lacked consensus, with no majority of more than \(\frac{2}{3}\) of the runs agreeing on the direction. 

\section{Figures}
Here are some figures that may clarify our intuitions and solutions presented in the paper.

\begin{figure*}
    \centering
    \includegraphics[width=\textwidth]{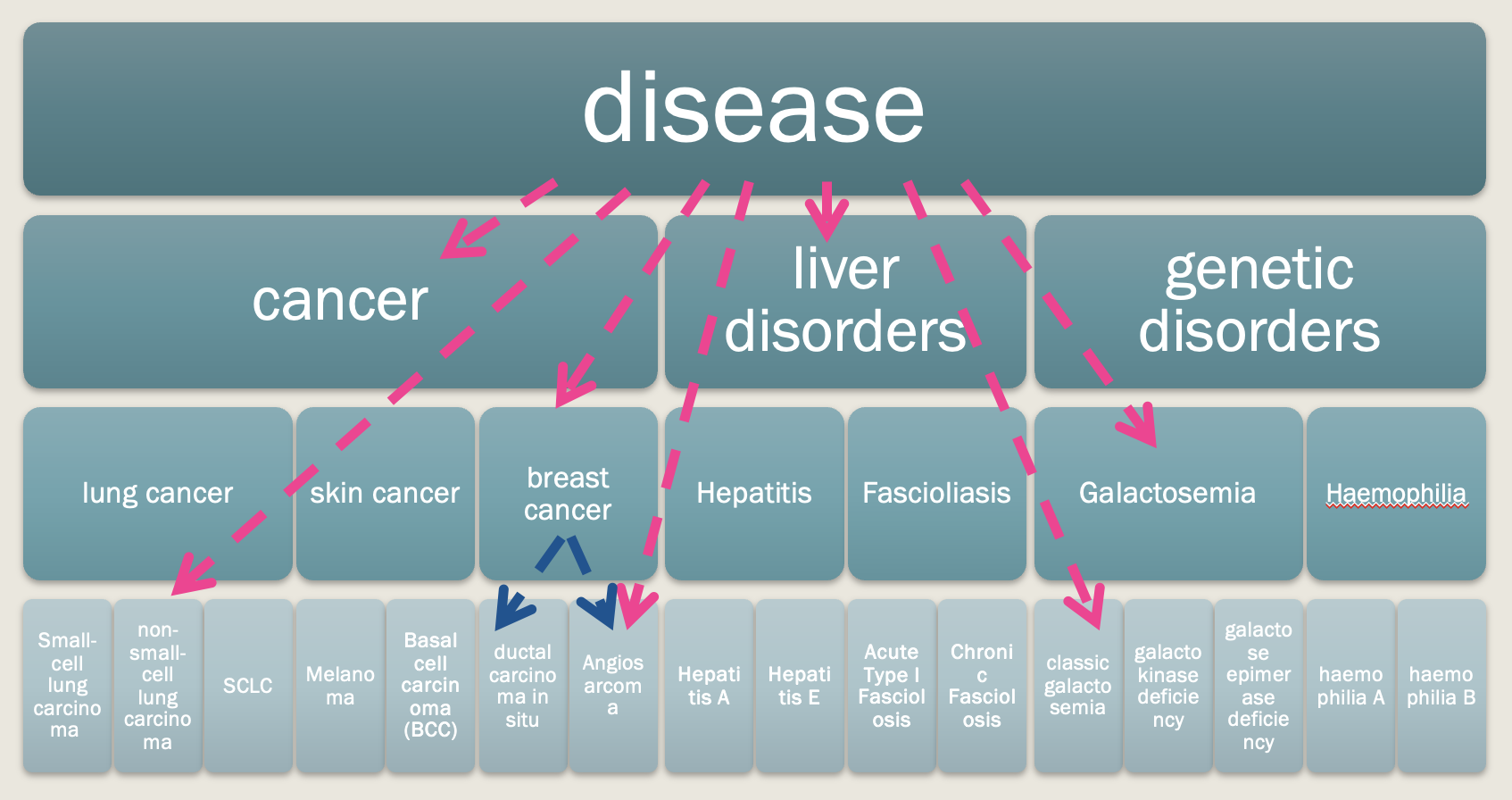} 

    \caption{\textbf{Co-occurrence behavior example} demonstrating why the more general the phrase, the more central the phrase. Each phrase can appear with any of the phrases that are more specific than it, making a phrase like "disease" a bridge between communities that is much more central than "breast cancer" in our graph.} 
    \label{betweenness_motivation_fig}
\end{figure*}

\clearpage

\begin{figure*}
     \centering
     \begin{subfigure}{\textwidth}
         \centering
         \includegraphics[width=\textwidth]{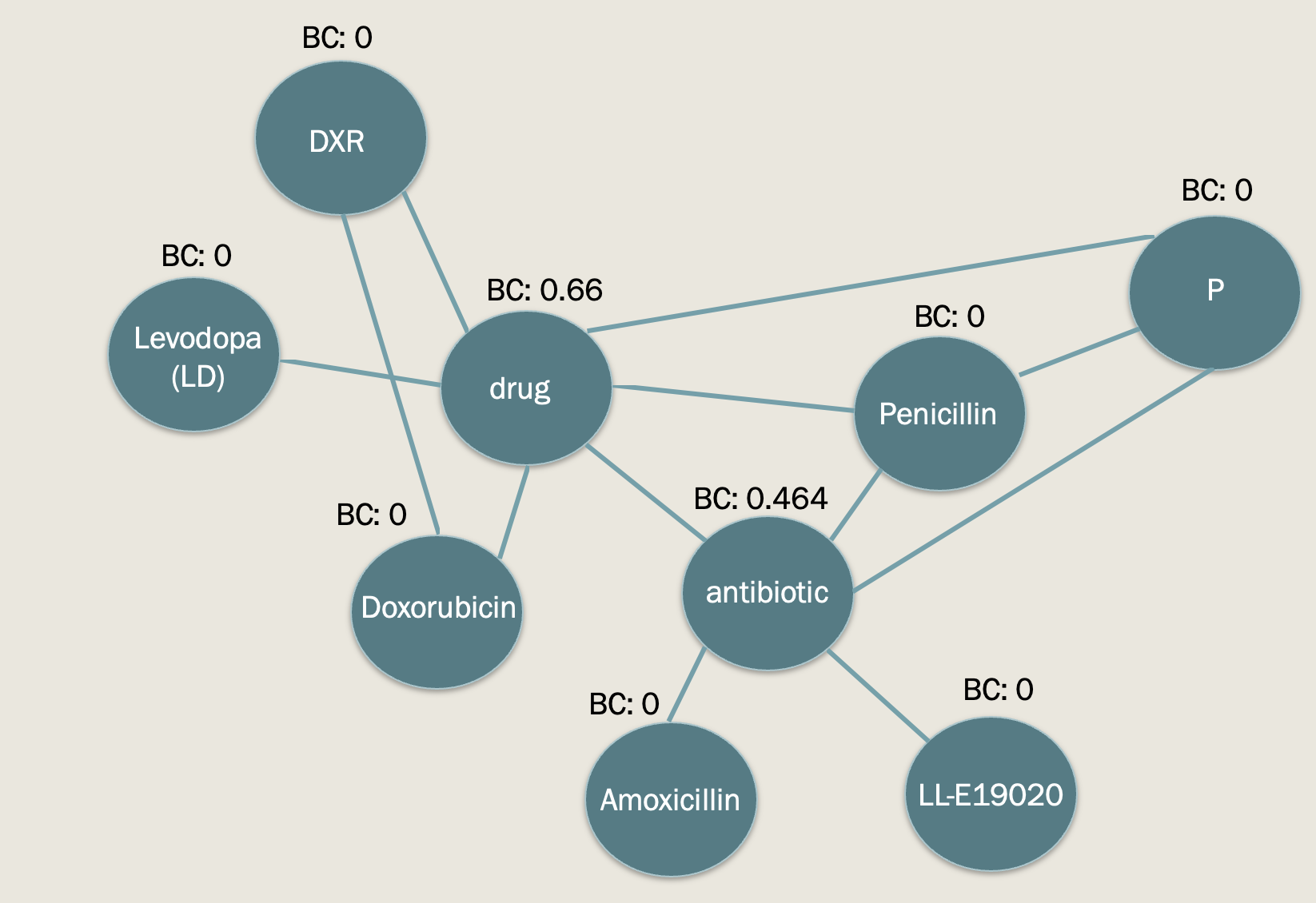}
     \end{subfigure}
     \begin{subfigure}{\textwidth}
         \centering
         \includegraphics[width=\textwidth]{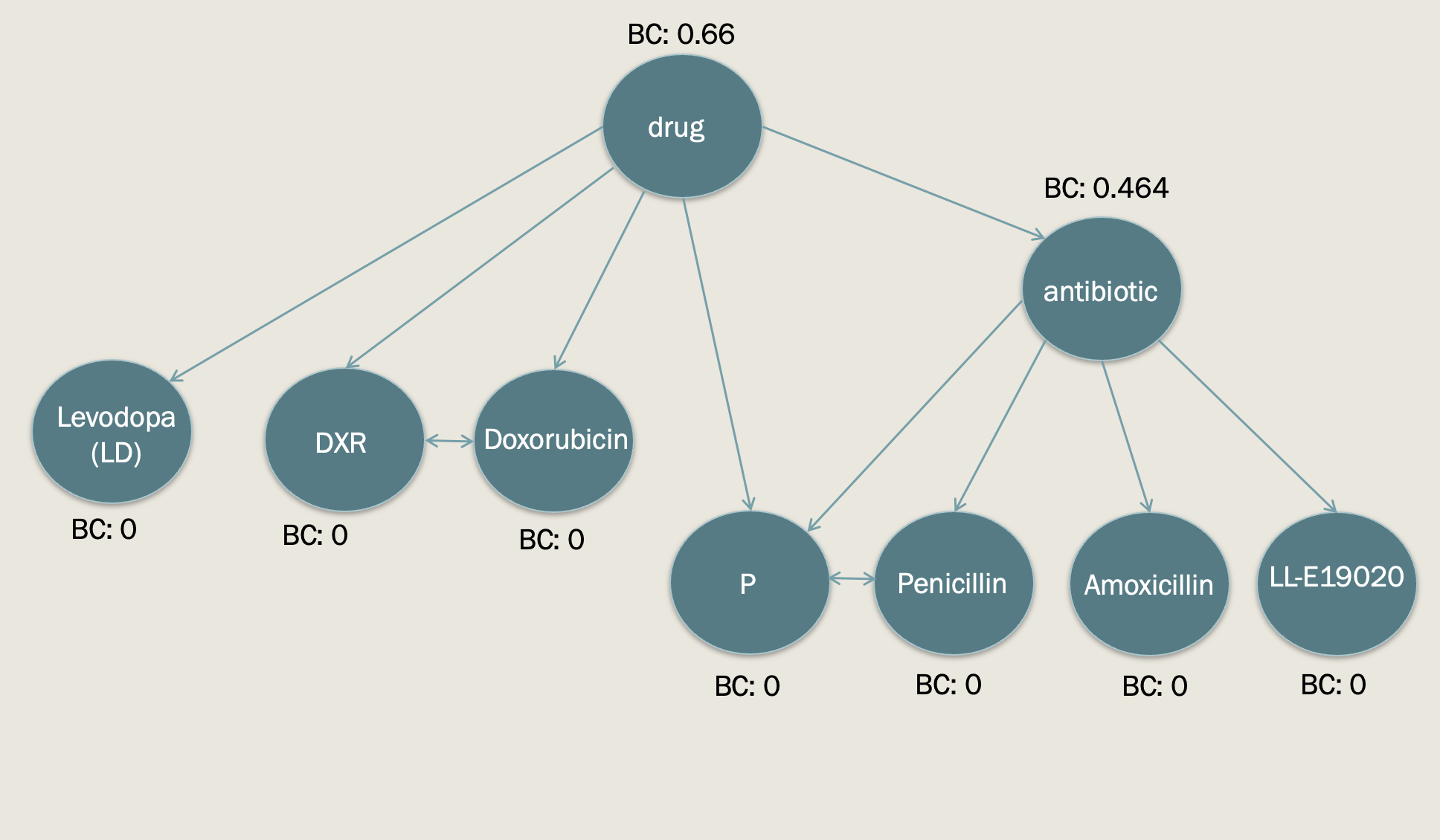}
     \end{subfigure}
     \caption{\textbf{Example of directions assignment} to the edges in the graph. The upper graph demonstrates the connections between phrases that appeared in our corpus, and their betweeness centrality (BC) values in this graph. The one below shows the result of a directed graph using them.}
\end{figure*}

     



\clearpage

\begin{figure*}
    \centering
    \includegraphics[width=\textwidth]{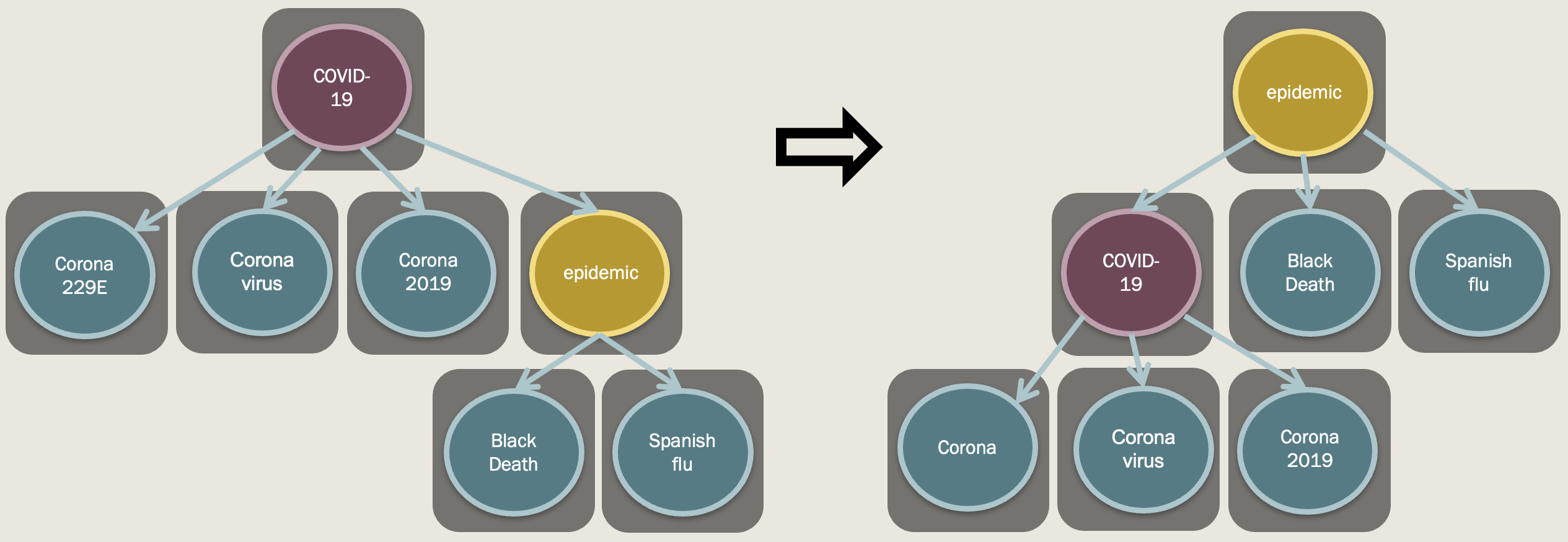}
    \caption{\textbf{Fixing Edge Direction} in cases where a name (e.g., "COVID-19") co-occurs with others in coreference chains more frequently than its general phrase neighbor (e.g., "epidemic"). Our solution (on the right) for correcting the directionality in these cases helps make the paths more accurate. (The gray background represents a concept)
}
    \label{fig_case1}
\end{figure*}

\begin{figure*}
    \centering
    \includegraphics[width=\textwidth]{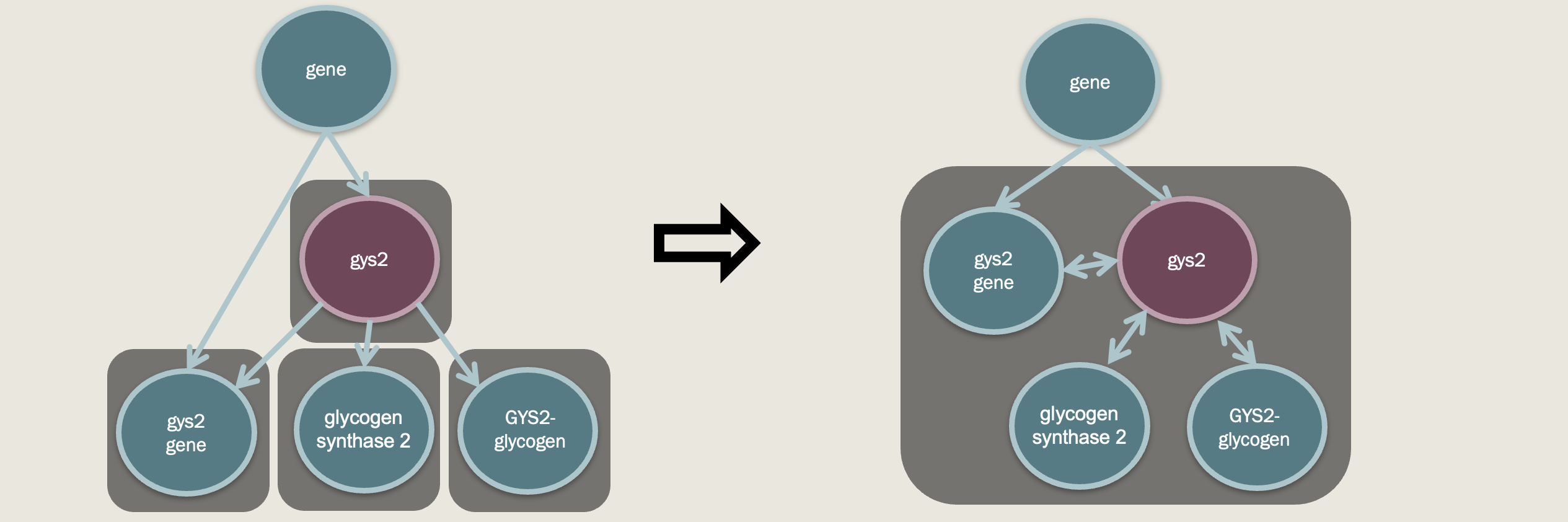}
    \caption{\textbf{Union Nodes to a Concept} when a common name (e.g., "gys2") is incorrectly identified as being more hierarchical than its alias neighbors. Our solution (on the right) that is based on semantic similarity representations helps in solving such cases. (The gray background represents a concept)
}
    \label{fig_case2}
\end{figure*}

\begin{figure*}
    \centering
    \includegraphics[width=\textwidth]{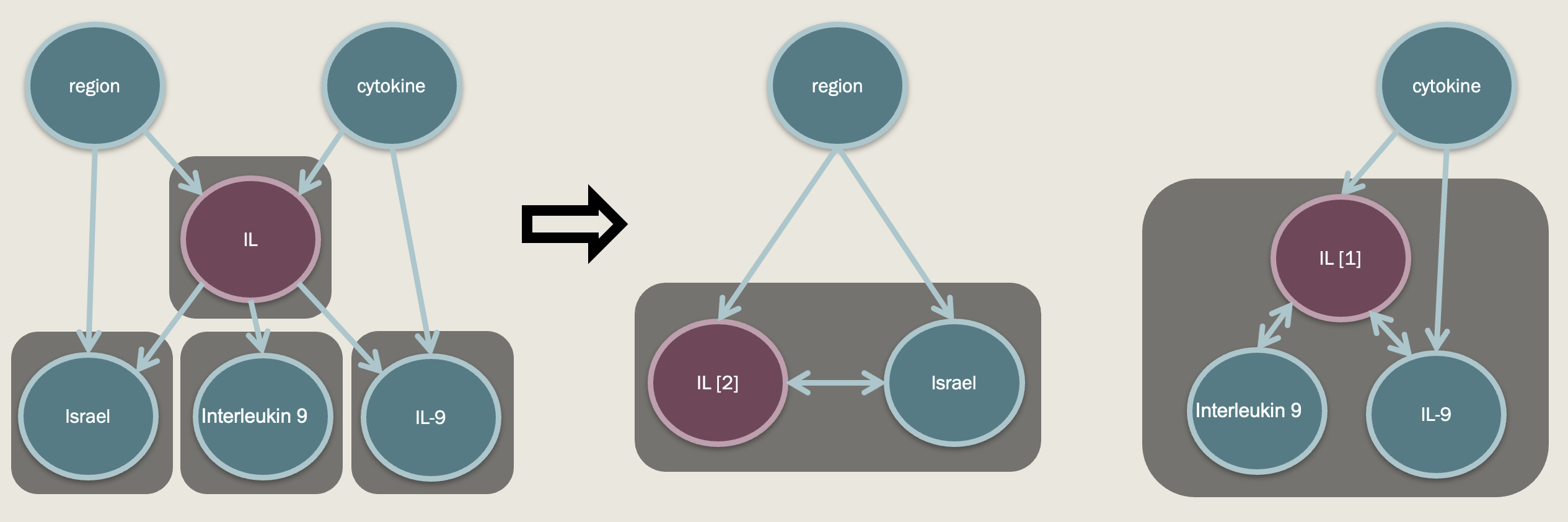}
    \caption{\textbf{Splitting Nodes with Multiple Senses} when an abbreviation (e.g., "IL") needs to be split into separate alias nodes for its different meanings (e.g., "Israel" and "IL-9"). Our solution is depicted on the right, showing the rearrangement of the subgraph into concepts, using the semantic similarity representations. (The gray background represents a concept)
}
    \label{fig_case3}
\end{figure*}

\end{document}